\documentclass{article}

\usepackage{microtype}
\usepackage{graphicx}
\usepackage{subfigure}
\usepackage{booktabs}

\usepackage{hyperref}

\usepackage[accepted]{icmlish}

\usepackage{macros}
\usepackage{outlines}
\usepackage[capitalise]{cleveref}

\begin{document}

\twocolumn[
\icmltitle{Ordering Dimensions with Nested Dropout Normalizing Flows}

\begin{icmlauthorlist}
\icmlauthor{Artur Bekasov}{ed}
\icmlauthor{Iain Murray}{ed}
\end{icmlauthorlist}

\icmlaffiliation{ed}{School of Informatics, University of Edinburgh, UK}

\icmlcorrespondingauthor{Artur Bekasov}{artur.bekasov@ed.ac.uk}

\icmlkeywords{Machine Learning, ICML}

\vskip 0.3in
]

\printAffiliationsAndNotice{} 

\begin{abstract}
The latent space of normalizing flows must be of the same dimensionality as their output space.
This constraint presents a problem if we want to learn low-dimensional, semantically meaningful representations.
Recent work has provided compact representations by fitting flows constrained to manifolds, but hasn't defined a density off that manifold.
In this work we consider flows with full support in data space, but with ordered latent variables. Like in PCA, the leading latent dimensions define a sequence of manifolds that lie close to the data.
We note a trade-off between the flow likelihood and the quality of the ordering, depending on the parameterization of the flow.
\end{abstract}

\section{Introduction}

Normalizing flows provide a way to parameterize complex distributions in high-dimensional spaces via a sequence of invertible transformations of a simple base distribution.
Unlike most other popular flexible generative models \citep[e.g.][]{kingma2013, goodfellow2014a_gan, du2019},
flows have a tractable likelihood, which simplifies training and model comparison. It is also easier to use tractable representations of distributions in other models and algorithms, such as for variational autoencoder (VAE) priors and posteriors.

However, generative models are not just used for estimating probabilities; we also fit these models to learn representations for down-stream tasks \citep{bengio2012}.
The latent spaces learned by normalizing flows can be semantically meaningful \citep{dinh2016_realnvp, kingma2018_glow} and can be useful for down-stream applications, such as classification \citep{nalisnick2019}.

For standard flows the latent dimensionality must match the input dimensionality, in order to satisfy the invertibility constraint.
This constraint presents a limitation for representation learning, where we are often interested in learning lower-dimensional representations (\ie manifolds) that capture high-level semantic concepts \citep{bengio2012}.

When data is constrained to a known manifold, it is possible to build a valid tractable flow on that space \citep{gemici2016,rezende2020}.
For data without strict constraints, \citet{kumar2020_injective_flow} and \citet{brehmer2020_mfmf} fit an invertible flow defined on a low-dimensional manifold, and bring the manifold close to the data by minimizing square error. Neither of these methods provide a tractable probabilistic model in the data space.

In this work we maintain a full-dimensional flow, but aim to order the latent variables, such that taking the first $K$ of them will reconstruct data with small square error. For a linear flow, our approach implements Principal Components Analysis (PCA)\@.
We encode this principle with an additional loss based on \emph{nested dropout} \citep[ND,][]{rippel2014_nested_dropout}.
We show that our models successfully learn low dimensional representations, while obtaining similar likelihoods to standard flows. However, we find that the parameterization of the flow influences the representations, and we must make a trade-off between the compactness of representations and the likelihood of the flow. We hope that these findings will motivate research into flows that more naturally represent low-dimensional structure.

\section{Nested dropout}

Nested dropout was introduced by \citet{rippel2014_nested_dropout} in the context of autoencoders, with a goal of imposing an ordering on representation dimensions.
For a linear autoencoder, nested dropout fits PCA\@.
In a non-linear case, the learned ordering was shown to be useful for efficient retrieval and adaptive compression.

As in dropout \citep{srivastava2014}: nested dropout randomly ``drops'' (\ie zeros out) representation dimensions during training.
However, rather than dropping features independently, we sample an index $k \sim p_k\br{\cdot}$, and drop dimensions $k+1 \dots K$, \ie all dimensions above that index, assuming $K$ is the latent dimensionality.

We define $\vz_{\downarrow k}$ as a representation $\vz$ where all the dimensions above $k$-th have been dropped, as described above.
We then define a low-dimensional reconstruction of a given datapoint $\vx$ for a given index $k$ as 
\begin{equation}
\label{eq:reconstruction}
\hat{\vx}_{\downarrow k} = g_\theta\br{f_\theta\br{\vx}_{\downarrow k}},
\end{equation}
where $f_\theta$ and $g_\theta$ are the encoder and the decoder respectively, with parameters $\theta$.

To train the autoencoder for a given dataset $\set{\vx_n}_{n=1}^N$, we minimize the following objective:
\begin{equation}
\label{eq:reconstruction_loss}
\mathcal{L}\br{\vect{\theta}} = \frac{1}{N}\sum_{n=1}^N \expect{d\br{\vx_n, \hat{\vx}_{n\downarrow k}}}{k\sim{p_k}},
\end{equation}
where 
$d$ is the chosen distance metric, typically $L_2$ distance.
This set-up encourages the auto-encoder to put more information into dimensions that correspond to smaller $k$ indices.

\citeauthor{rippel2014_nested_dropout} used a geometric distribution for $p_k$, traditionally parameterized as $p_k\br{k} \!=\! \br{1 - p}^{k-1}p$ for a given Bernoulli probability $p$.
We follow their choice in this work.

\subsection{Normalizing flows with nested dropout}
\label{sec:nested_dropout_flows}

We start with the standard training objective of a normalizing flow.
For a given parametric invertible function $f_{\vect{\theta}}$ and a base density $\pi$, the density of a given datapoint $\vx$ is:
\begin{equation}
\label{eq:flow_density}
p_\vect{\theta}\br{\vx} = \pi\br{f_{\vect{\theta}}\br{\vx}}\abs{\det\br{\deriv{f_{\vect{\theta}}}{\vx}}}.
\end{equation}
Given a dataset $\set{\vx_n}_{n=1}^N$, we fit the flow parameters $\vect{\theta}$ by minimizing the negative log likelihood objective:
\begin{equation}
    \label{eq:likelihood_loss}
    \mathcal{L}\br{\vect{\theta}} = -\frac{1}{N} \sum_{n=1}^N \log p_\vect{\theta}\br{\vect{x}_n}.
\end{equation}
We can redefine a lower-dimensional reconstruction, \cref{eq:reconstruction}, for a normalizing flow by treating the invertible function $f_\theta$ as the ``encoder'', and its inverse $f^{-1}_\theta$ as the ``decoder'':
\begin{equation}
    \label{eq:flow_reconstruction}
    \tilde{\vx}_{\downarrow k} = f_\theta^{-1}\br{f_\theta\br{\vx}_{\downarrow k}}.
\end{equation}
We combine \cref{eq:reconstruction_loss,eq:flow_reconstruction,eq:likelihood_loss} into a single objective, which specifies that the flow should have high likelihood, while also giving good reconstructions from low-dimensional parts of its representation:
\begin{equation}
    \textstyle
    \mathcal{L}\br{\vect{\theta}} = {\frac{1}{N}}\!\sum_{n=1}^N\bigl(
    -\log p_\vect{\theta}\br{{\vect{x}_n}}
    + \lambda\,\expect{d\br{\vx_n, \tilde{\vx}_{n\downarrow k}}}{k\sim{p_k}}\bigl),
\end{equation}
where $\lambda$ is a hyper-parameter that balances the two losses.
In line with \citeauthor{rippel2014_nested_dropout} we estimate the expectation with a single Monte-Carlo sample.

The loss is similar to the one used by \citet[][Eq.\;21]{brehmer2020_mfmf}, with two distinctions.
First, $k$ is sampled from $p_k$ for each datapoint, rather than pre-set according to the desired manifold dimensionality.
Second, while \citeauthor{brehmer2020_mfmf} use two separate flows, one for each part of the objective, we train a single flow for both.

\section{Experiments}

\subsection{Synthetic dataset}

\begin{table}[t]
    \vskip -0.07in
    \caption{Average test log-likelihood (in nats) and reconstruction MSE when projecting down to 1 or 2 dimensions for the synthetic dataset. Mean $\pm$ 2 standard deviations over 10 different initializations. Superscript$^*$ for models trained with nested dropout.}
    \label{tab:synthetic_results}
    \vskip 0.1in
    \begin{center}
    \begin{small}
    \begin{sc}
    \bgroup
    \setlength{\tabcolsep}{0.5em}
    \begin{tabular}{lccc}
    \toprule
    & LL & MSE(2) & MSE(1) \\ \midrule
    PCA & $\hphantom{-0.00}$ -- $\hphantom{{}\pm 0.000}$ & $0.003 \hphantom{{}\pm 0.000}$ & $0.037\hphantom{{}\pm 0.000}$\\
    \midrule
    QR & $-0.804 \pm 0.000$  & $0.240 \pm 0.053$ & $0.321 \pm 0.025$ \\
    QR$^*$ & $-0.804 \pm 0.000$ & $0.003 \pm 0.000$ & $0.037 \pm 0.000$\\
    \midrule
    LU & $-0.804 \pm 0.000$ & $0.051 \pm 0.042$ & $0.257 \pm 0.130$\\
    LU$^*$ & $-0.805 \pm 0.001$ & $0.008 \pm 0.007$ & $0.037 \pm 0.001$\\
    \bottomrule
    \end{tabular}
    \egroup
    \end{sc}
    \end{small}
    \end{center}
    \vskip -0.1in
\end{table}

As a simple testcase, we sample data
from a centered 3-dimensional normal distribution that is scaled along the axis and then rotated.
The eigenvalues of the covariance are $1$, $0.1$, and $0.01$, so the target distribution looks like a fuzzy elliptical disc embedded into 3 dimensions.
We sample $10^4$ points for training, and another $10^4$ points for evaluation.

We use a simple flow with a standard normal base distribution and a single invertible linear transformation.
Such a flow can express the target distribution by learning to scale/rotate the base distribution as necessary.
However, due to the spherical symmetry of the base distribution, the likelihood is invariant to permuting the dimensions (or any orthogonal transformation) of the base distribution.

\begin{figure}
    \begin{center}
    \centerline{\includegraphics[width=0.8\columnwidth]{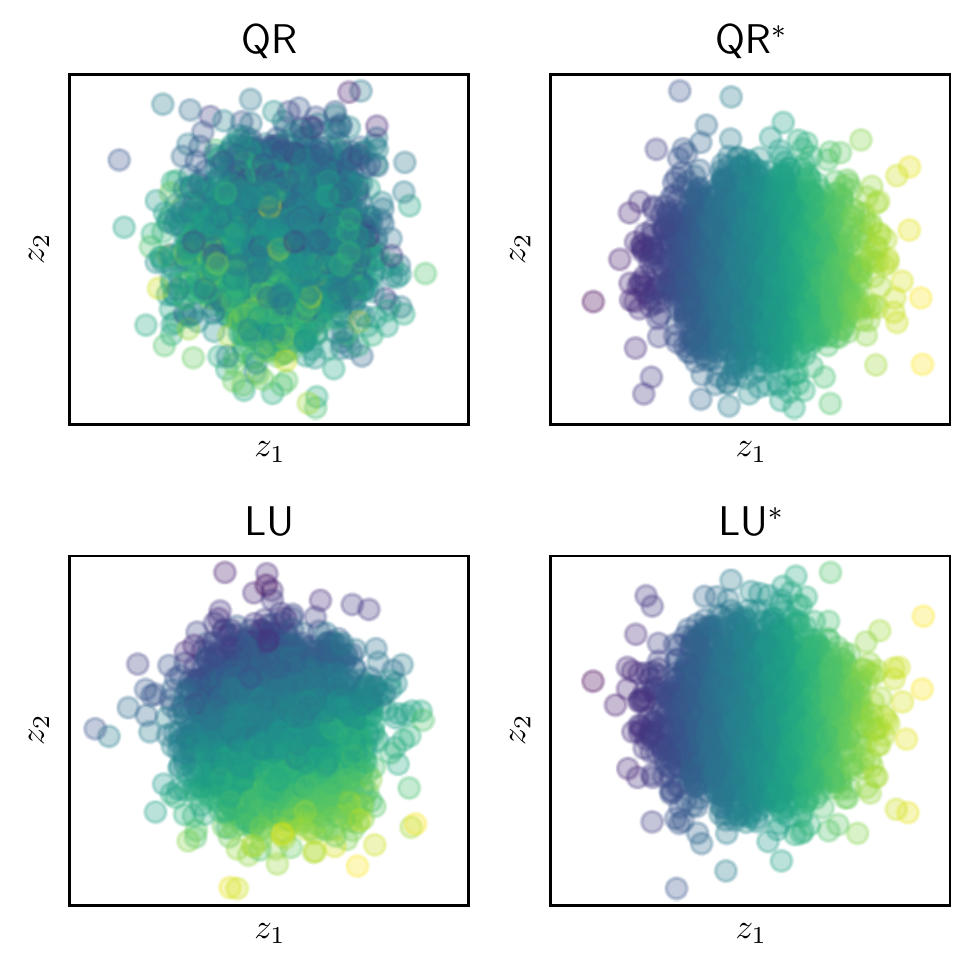}}
    \vskip -0.2in
    \caption{2D projections $\vz$ of the 3D synthetic dataset%
    , colored by the value of the first principal component of PCA. Superscript$^*$ for models trained with nested dropout.}
    \label{fig:synthetic_proj}
    \end{center}
    \vskip -0.2in
\end{figure}

The invertible linear transformation is parameterized by either an LU decomposition with a random, fixed permutation matrix $\mat{P}$ \citep{kingma2018_glow}, or a QR decomposition \citep{hoogeboom2019} with the orthogonal matrix parameterized
by 3 
Householder transformations \citep{tomczak2016_householder}.
We train the flow using the Adam optimizer \citep{kingma2014} for $30{\times}10^3$ iterations with a batch size of 500.
We set the reconstruction coefficient to $\lambda=20$ for nested dropout runs, with $p_k$ set to a geometric distribution with $p=0.33$. 

We use PCA as a baseline for reconstruction results, where we project onto 1 or 2 principal components. Likelihood and reconstruction results are summarized in \cref{tab:synthetic_results}. We visualise some of the projections in \cref{fig:synthetic_proj}.

For both parameterizations the additional loss improves reconstructions from 1 or 2 dimensions.
For the QR parameterization, matching PCA's optimal representation isn't a restriction, and the test likelihood is maintained.
The results for the LU parameterization have higher variance with nested dropout.
The permutation matrix $\mat{P}$ is not learned, and its initialization matters.

\subsection{Images}

To evaluate the method on high-dimensional data, we fit a normalizing flow with nested dropout to Fashion-MNIST images \citep{xiao2017_fashion_mnist}.
To simplify the model architecture, we pad the images by 2~pixels on each side, giving $32{\times}32$ or 1024-dimensional images.
We use the provided test set, but split the provided training set into $50\,000$ training images and $10\,000$ validation images.
We dequantize the images by adding uniform noise $\mat{U} \in [0,1)^{32{\times}32}$.

We use a RQ-NSF (C) image flow
\citep{durkan2019_nsf} containing
3 multi-scale levels with 4 transformation steps per level, where each step consists of an activation normalization layer, a 1x1 convolution, and a rational-quadratic coupling transform.
Residual convolutional networks with 3 blocks and 128 hidden channels parameterize the 4-bin rational-quadratic splines in the coupling transforms.
We train all models for $100{\times}10^3$ iterations, with a batch size of 256.
We anneal the learning rate of the Adam optimizer from $5{\times}10^{-4}$ down to zero according to a cosine schedule \citep[Eq. 5]{loshchilov2016}.

\begin{figure}[t]
    \begin{center}
    \centerline{\includegraphics[width=\columnwidth]{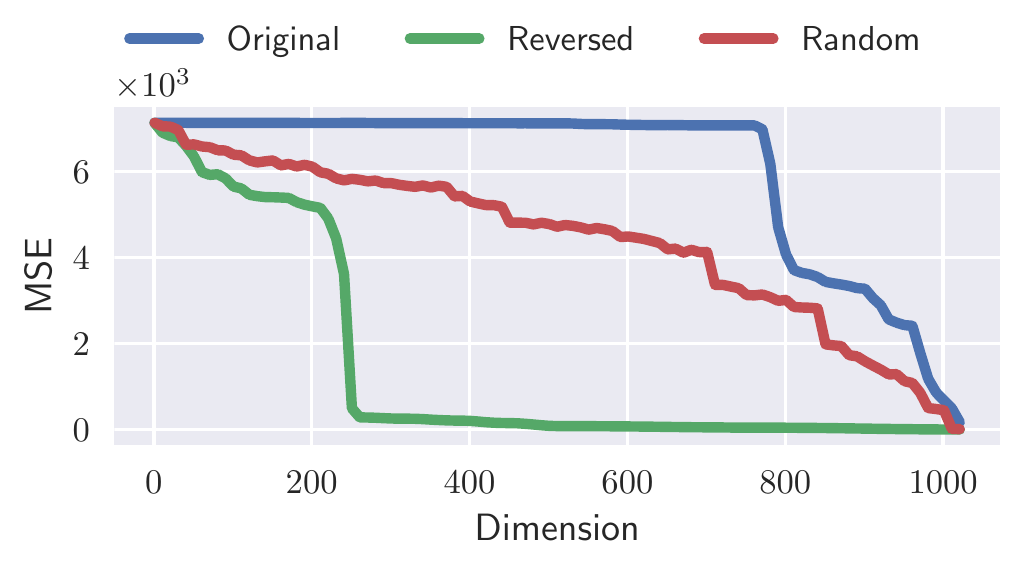}}
    \vskip -0.15in
    \caption{Mean squared error of Fashion-MNIST reconstructions for RQ-NSF (C) flow against the number of retained dimensions, varying the order in which the dimensions are dropped.}
    \label{fig:nd_order}
    \end{center}
    \vskip -0.2in
\end{figure}

\begin{figure}[t]
    \begin{center}
    \centerline{\includegraphics[width=\columnwidth]{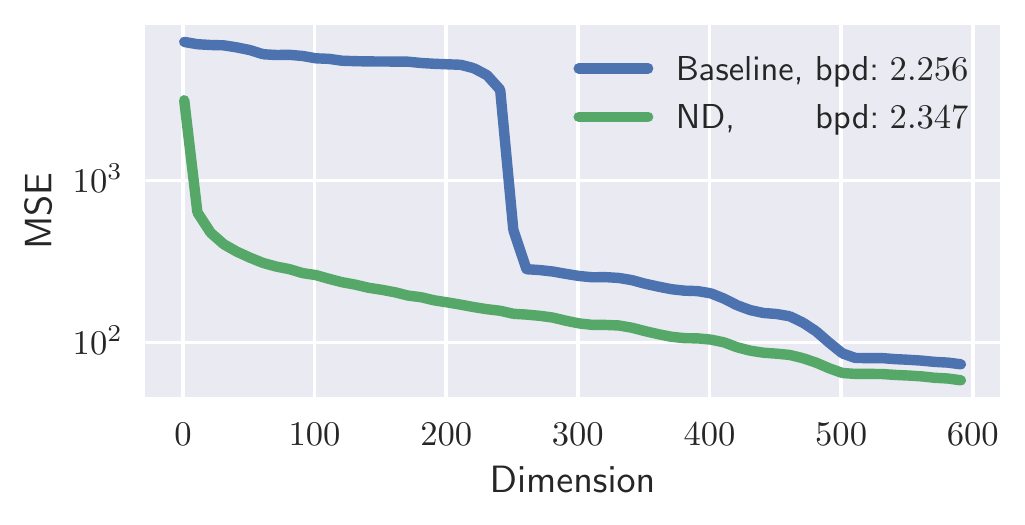}}
    \vskip -0.15in
    \caption{Mean squared error of Fashion-MNIST reconstructions for RQ-NSF (C) flow against the number of retained dimensions. Comparing a flow trained with nested dropout to the baseline without the additional loss. \emph{bpd} is the negative test log-likelihood in bits per dimension (lower is better).}
    \label{fig:baseline_vs_nd}
    \end{center}
    \vskip -0.2in
\end{figure}

\begin{figure}[ht]
    \begin{center}
    \bgroup
    \setlength{\tabcolsep}{0.1em}
    \begin{small}
    \begin{sc}
    \def\arraystretch{0.5}%
    \begin{tabular}{cc}
        \includegraphics[width=.49\columnwidth]{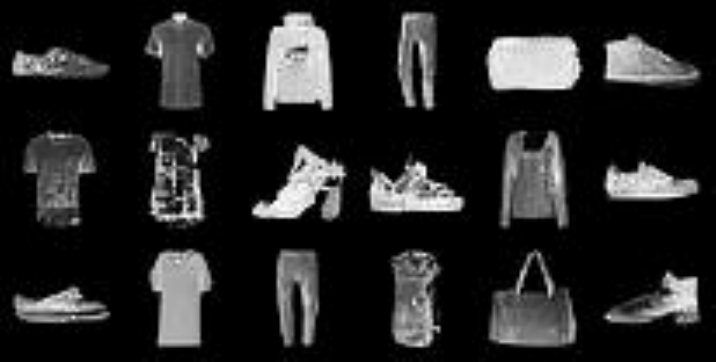} &
        \includegraphics[width=.49\columnwidth]{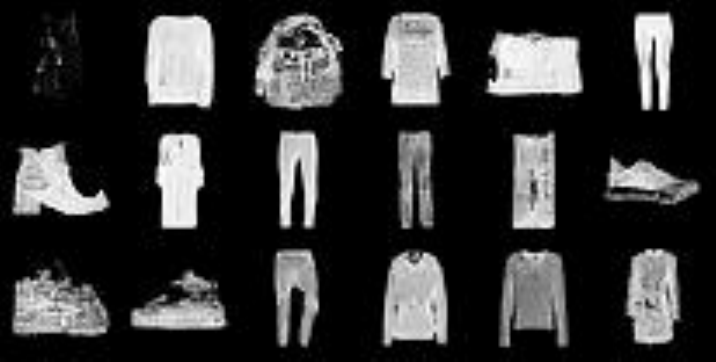} \\ 
        \includegraphics[width=.49\columnwidth]{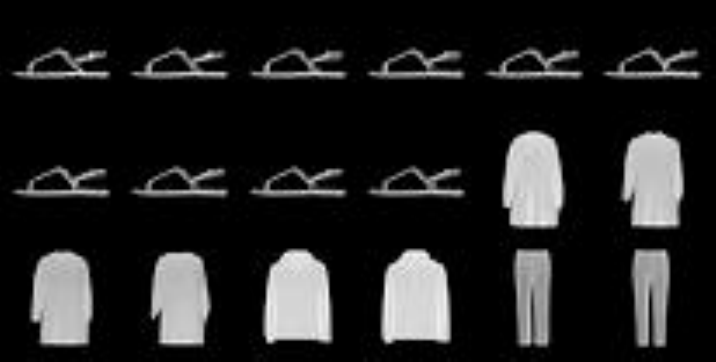} &
        \includegraphics[width=.49\columnwidth]{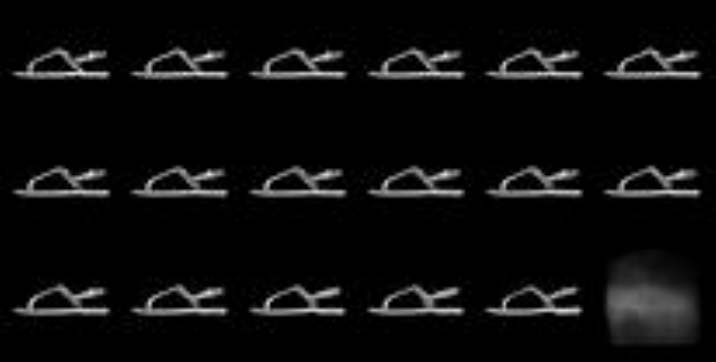}
    \end{tabular}
    \end{sc}
    \end{small}
    \egroup
    \vskip -0.1in
    \caption{\textbf{Top:} samples from the flow. \textbf{Bottom:} reconstructions for an input image in the top left, decaying the latent dimension from 510 to 1 (left-to-right, top-to-bottom). \textbf{Left}: RQ-NSF (C) baseline. \textbf{Right}: RQ-NSF (C) with nested dropout.}
    \label{fig:samples_reconstructions}
    \end{center}
    \vskip -0.1in
\end{figure}

\subsubsection{Ordering in a multi-scale architecture}

The multi-scale architecture with variable splitting, introduced by \citet{dinh2016_realnvp} and used in RQ-NSF (C), {already} induces a partial ordering on variables, even without the additional nested dropout objective.
Some variables undergo more transformations than others, and those undergoing more transformations are eventually transformed at ``coarser scale'', \ie by transformations conditioned on variables that are further away spatially.
\citet[Appendix D]{dinh2016_realnvp} demonstrate that such variables contain higher-level semantic information.

\cref{fig:nd_order} demonstrates the effect of this ordering.
There is a sudden drop in error after including ${\approx}250$ dimensions from the reversed order. In our implementation of the multi-scale transform, it is this last quarter of the variables that are transformed at all 3 multi-scale levels.
We use the reversed order in all further experiments, both during training with nested dropout, and during evaluation. 

\subsubsection{Training with nested dropout}

We now try to understand whether we can use nested dropout to improve upon the baseline RQ-NSF (C) model results in terms of reconstruction accuracy.
We set the reconstruction coefficient $\lambda = 10^{-3}$, and we set $p = 10^{-3}$ for the geometric distribution $p_k$.
Reconstruction and likelihood results are shown in \cref{fig:baseline_vs_nd}.

We note that the reconstruction curve is drastically improved, in particular for ${<}\,250$ dimension projections.
However, we also note that the test log-likelihood of the model suffers, being reduced by $\approx\,$4\% as a result of applying nested dropout.
We hypothesize that this trade-off is controlled by the values of $\lambda$ and $p$, which we explore further in \cref{sec:hyperparams}.

We show samples in the ambient space and reconstructions for a baseline flow and a nested dropout flow in \cref{fig:samples_reconstructions}.
Even though the likelihoods of the two models differ, their samples are of comparable perceptual quality.
It has been noted that the relationship between the perceptual quality of samples and likelihood is complex \citep{theis2015}.
The reconstructions demonstrate that nested dropout indeed guides the flow towards ordering the representations.

\begin{figure}[t]
    \begin{center}
    \centerline{\includegraphics[width=0.99\columnwidth]{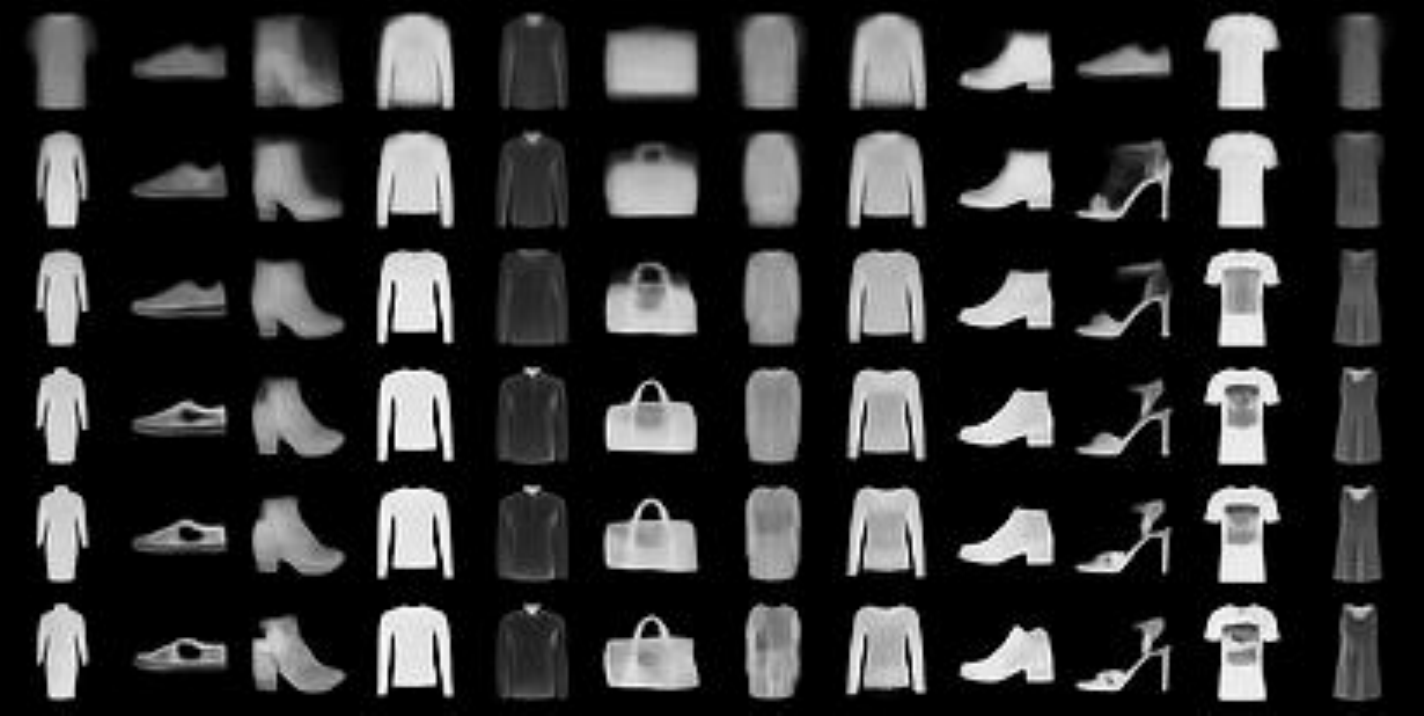}}
    \vskip -0.05in
    \caption{Samples from RQ-NSF (C) trained with nested dropout. Each row shows samples with a different number of latent dimensions retained: $2,4,8,16,32,64$ for each row in order.}
    \label{fig:manifold_samples}
    \end{center}
    \vskip -0.30in
\end{figure}

For the nested dropout flow, \cref{fig:manifold_samples} shows samples from the flow varying the number of dropped dimensions.%
\footnote{As noted by \citet[Section 2.C,][]{brehmer2020_mfmf}, sampling in this fashion does not correspond to sampling from densities on corresponding manifolds.}
Using 2 dimensions gives a generic, blurry item, which becomes sharper and more detailed given more latent features.

\subsubsection{Hyper-parameters}
\label{sec:hyperparams}

\begin{figure}[t]
    \vskip -0.07in
    \begin{center}
    \centerline{\includegraphics[width=\columnwidth]{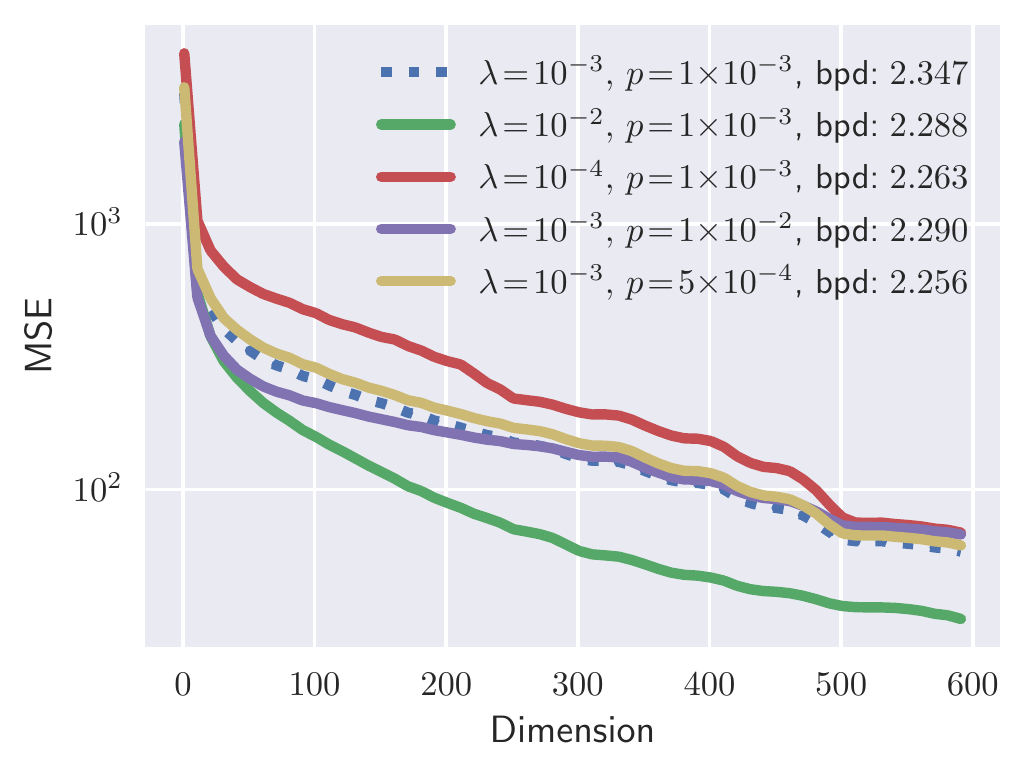}}
    \vskip -0.1in
    \caption{Mean squared error of Fashion-MNIST reconstructions for RQ-NSF (C) flow trained with nested dropout against the number of retained dimensions, varying hyper-parameters $\lambda$ and $p$. The dotted line for the baseline nested dropout model. \emph{bpd} is the negative test log-likelihood in bits per dimension (lower is better).}
    \label{fig:hyperparams}
    \end{center}
    \vskip -0.25in
\end{figure}

We perform a limited grid-search for the $\lambda$ and $p$ hyper-parameters.
We perturb each hyper-parameter independently, starting with the baseline values used in the previous section. 
Results are shown in \cref{fig:hyperparams}.

The value of $p$ has a limited effect on the reconstruction curve.
A lower value (which causes more dimensions to be dropped on average during the run) marginally improves the MSE numbers for $<200$ latent dimensions.

The effect of changing $\lambda$ is more pronounced.
Larger values cause a noticeable improvement in reconstruction results, while lower values have the opposite effect.

All 4 models improve upon the baseline in terms of test log-likelihood, which is surprising. There could be an interaction between $\lambda$ and $p$ hyper-parameters, or large variation across runs.
Unfortunately, all but one result, including the one with the best reconstruction results, have slightly lower likelihoods than the standard flow.

\section{Discussion}

Nested dropout is a simple way to encourage a flow to represent data as closely
as possible in the top, \emph{principal} elements of its latent representation.
Given the large redundancy in the way flows parameterize distributions, we can
choose flows with similar likelihoods that provide much better low dimensional
representations.

The small reduction in likelihood suggests
that existing flows have some inductive bias \emph{against} interpretable latent spaces.
We also can't always apply nested dropout, because
the flow must be evaluated in both directions on every iteration.
For architectures with one-pass sampling, training is typically slower by a factor of two.
Flows without a cheap inverse, \eg auto-regressive flows, would be impractical.

If data really lies on a low-dimensional manifold, densities in data space are
not well defined, and other work is more appropriate \citep{kumar2020_injective_flow,brehmer2020_mfmf}.
However, PCA is often applied when data aren't really
restricted to a subspace of fixed dimensionality.
We hope that nested dropout flows will be useful in similar circumstances.

\section*{Acknowledgements}

This work was supported in part by the EPSRC Centre for Doctoral Training in Data Science, funded by the UK Engineering and Physical Sciences Research Council (grant EP/L016427/1) and the University of Edinburgh.

\bibliography{bibliography.bib}
\bibliographystyle{icml2020}

\end{document}